\documentclass{article}

\usepackage[nonatbib,final]{neurips_2022_ml4ad}

\usepackage[utf8]{inputenc} 
\usepackage[T1]{fontenc}    
\usepackage{url}            
\usepackage{booktabs}       
\usepackage{amsfonts}       
\usepackage{nicefrac}       
\usepackage{microtype}      
\usepackage{xcolor}         

\usepackage{epsfig}
\usepackage{graphicx}
\usepackage{amsmath}
\usepackage{amssymb}

\usepackage{chngcntr}
\usepackage[export]{adjustbox}
\usepackage{caption}
\usepackage{subcaption}
\usepackage{diagbox}
\usepackage{enumitem}
\setlist{nosep}
\usepackage{comment}
\usepackage{multirow}
\usepackage{makecell}
\usepackage{placeins}
\usepackage{float}
\newcommand\dunderline[3][-1pt]{{%
  \sbox0{#3}%
  \ooalign{\copy0\cr\rule[\dimexpr#1-#2\relax]{\wd0}{#2}}}}
\newcommand{\uu}[1]{\dunderline{1pt}{#1}}

\usepackage[title]{appendix}
\usepackage{wrapfig}
\usepackage{xspace}

\makeatletter
\DeclareRobustCommand\onedot{\futurelet\@let@token\@onedot}
\def\@onedot{\ifx\@let@token.\else.\null\fi\xspace}

\def\eg{\emph{e.g}\onedot} 
\def\ie{\emph{i.e}\onedot}

\def\etal{\emph{et al}\onedot}
\makeatother

\usepackage[capitalize]{cleveref}
\crefname{section}{Sec.}{Secs.}
\Crefname{section}{Section}{Sections}
\Crefname{table}{Table}{Tables}
\crefname{table}{Tab.}{Tabs.}

\title{Stress-Testing Point Cloud Registration on Automotive LiDAR}

\author{%
  Amnon Drory    \\
  \And
  Raja Giryes \\
  \\
  Tel-Aviv University \\
  \And
  Shai Avidan \\
}

\begin{document}

\maketitle

\begin{abstract}

Rigid Point Cloud Registration (PCR) algorithms aim to estimate the 6-DOF relative motion between two point clouds, which is important in various fields, including autonomous driving. Recent years have seen a significant improvement in \emph{global} PCR algorithms, \ie algorithms that can handle a large relative motion. This has been demonstrated in various scenarios, including indoor scenes, but has only been minimally tested in the Automotive setting, where point clouds are produced by vehicle-mounted LiDAR sensors. In this work, we aim to answer questions that are important for automotive applications, including: which of the new algorithms is the most accurate, and which is fastest? How transferable are deep-learning approaches, \eg what happens when you train a network with data from Boston, and run it in a vehicle in Singapore? How small can the overlap between point clouds be before the algorithms start to deteriorate? To what extent are the algorithms rotation invariant? 
Our results are at times surprising. When comparing robust parameter estimation methods for registration, we find that the fastest and most accurate is not one of the newest approaches. Instead, it is a modern variant of the well known RANSAC technique. We also suggest a new outlier filtering method, Grid-Prioritized Filtering (GPF), to further improve it.    
An additional contribution of this work is an algorithm for selecting challenging sets of frame-pairs from automotive LiDAR datasets. This enables meaningful benchmarking in the Automotive LiDAR setting, and can also improve training for learning algorithms. 

We share our code and registration sets.\footnote{\url{https://github.com/AmnonDrory/LidarRegistration}}

\end{abstract}


\section{Introduction}
\label{sec:intro}

Rigid Point Cloud Registration (PCR) is an important task in many fields, including autonomous driving. Its goal is to estimate the relative motion between two point clouds, in 6 degrees of freedom (x,y,z translation; pitch, roll and yaw). Recent years have seen a significant improvement in \emph{global} PCR algorithms, which can handle large relative motions. This is to a large extent thanks to deep-learned local features.
\begin{figure}[h!]
\begin{center}
\includegraphics[width=1\textwidth]{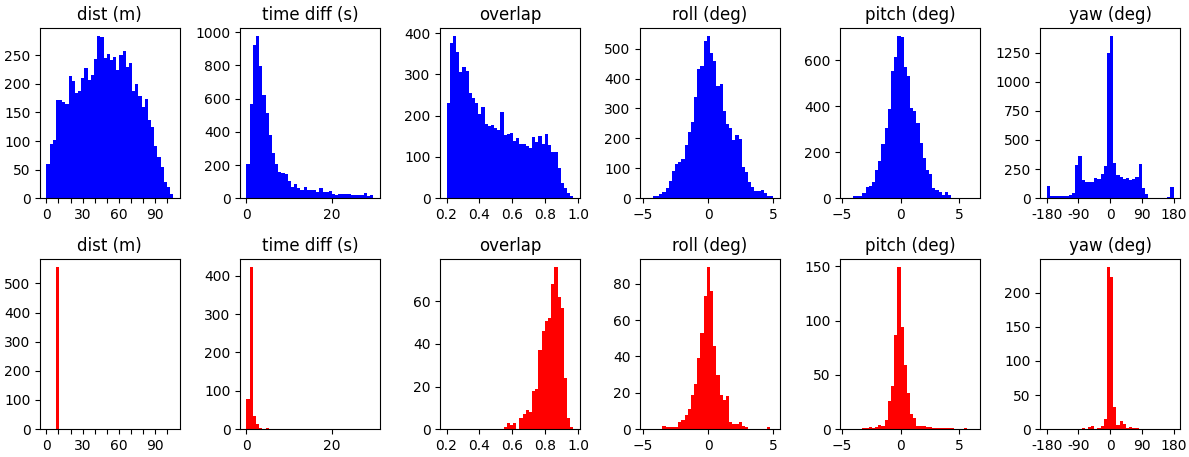}
\end{center}
   \caption{\textbf{Challenging registration sets.} The KITTI-10m registration set has been used extensively to evaluate PCR algorithms in the automotive LiDAR setting. However, it has recently become saturated: multiple algorithms achieve essentially perfect success. We suggest a new algorithm to select much more challenging registration sets. To compare the sets, we present here the distribution of test samples in \emph{KITTI-10m} (bottom) and our suggested \emph{Apollo-Southbay-Balanced} (top). We show the distribution of samples by (from left to right): distance between the pair of point clouds, time offset, overlap between scans, and rotation in three separate axes. \emph{Apollo-Southbay-Balanced} includes a balanced representation of all the relative motions that are encountered in a real driving scenario. It is much more challenging, containing significantly larger rotations and smaller overlaps than \emph{KITTI-10m}.}
\vspace{-0.04in}
\label{fig:set_stats}
\end{figure}
A popular and successful approach to global PCR is to divide it into two main stages: \emph{feature matching} and {robust parameter estimation}. In feature matching, a local descriptor (a.k.a feature) is calculated for each point in both clouds, and then each point in the source cloud is matched to the point in the target cloud that has the most similar descriptor to it. This results in a set of point pair matches, from which one could estimate a 6-DOF motion. However, the set of pairs contains many \emph{outliers}, \ie pairs whose relative motion does not follow the general motion between the two point clouds. This can be caused by independently moving objects in the scene, \eg vehicles or pedestrians. It can also be caused by incorrectly matched pairs, due to various causes including occlusions, partial overlap between point clouds, limitations of the local descriptors, and more. To overcome outliers, it is necessary to use robust parameter estimation algorithms for the 6-DOF motion. These include explicit filtering of outliers, RANSAC, and many other approaches.  

Recently, a dramatic improvement in the accuracy of PCR was achieved thanks to new, deep-learning based, features. Fully Convolutional Geometric Features (FCGF~\cite{FCGF}) have achieved dramatically improved accuracy compared to classical hand-crafted features such as FPFH~\cite{FPFH}.

New robust parameter estimation algorithms for the 6-DOF motion have also been suggested recently. These include DGR~\cite{DGR} and PointDSC~\cite{PointDSC}, which are based on deep learning, and also TEASER++~\cite{TEASER}, which is not learning-based. These techniques have been validated mostly on indoor scenes, showing a considerable success. When combined with weaker features (\eg FPFH), they show an ability to improve the results considerably. When combined with stronger features (\eg FCGF) they achieve state-of-the-art results. 

\begin{wrapfigure}{r}{0.5\linewidth}
\begin{center}
\begin{tabular}{c}
\includegraphics[width=0.5\columnwidth]{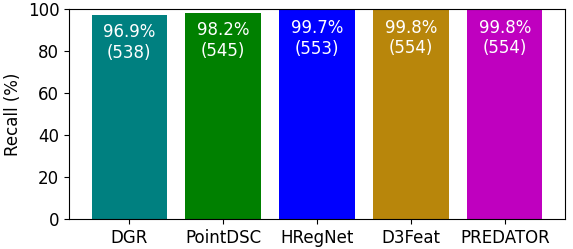}
\end{tabular}
\end{center}
\caption{\textbf{Saturation of KITTI-10m}. KITTI-10m has been the standard benchmark for LiDAR registration for the last few years. It is essentially saturated: several recent algorithms have achieved almost perfect recall on it, failing only on a handful of the 555 point-cloud pairs in its test set. The values shown here were taken from the corresponding papers \cite{DGR,PointDSC,HRegNet,D3Feat,PREDATOR}. This saturation happens also in other existing datasets for LiDAR registration.}
\label{fig:kitti_saturation}
\end{wrapfigure}

In the automotive field, point clouds are produced by vehicle mounted LiDAR sensors, and PCR is useful for estimating the ego-vehicular motion. There are several significant differences between automotive LiDAR PCR and other settings (such as single object scans and indoor): the scans are significantly larger, and often contain many distinct independent movers (vehicles, pedestrians, etc). In recent works, testing in the automotive LiDAR setting has been done almost entirely using the KITTI-10m registration set. This consists of pairs of frames from the KITTI Odometry dataset, that are separated by 10 meters. As a result, it appears that its test set consists almost entirely of high-overlap and small rotation situations. Thus, it fails to present a significant challenge to recent PCR algorithms, as many of them are able to achieve essentially perfect recall (up to a handful of failures at most, see \cref{fig:kitti_saturation}).

In this work, we aim at thoroughly testing these new and promising PCR approaches in the automotive LiDAR setting. Our first goal is to achieve a meaningful ranking between these algorithms. We are interested not only in accuracy, but also in running time, which is critical for realistic applications. To get a meaningful comparison, we need a better benchmark than the ones previously used. We achieve this by using larger and more modern LiDAR datasets, and also by selecting a much more challenging set of frame-pairs (see \cref{fig:set_stats}). We developed an algorithm that selects a set of frame pairs that is balanced in the sense that it contains various relative motions (large and small rotations, short and large offsets, etc), a range of overlap ratios, and a fair sampling from each of the driving sequences in a given dataset. Using these challenging sets, we get an insight into questions such as how low the overlap can get before registration results significantly deteriorate, to what extent the local descriptors are rotation invariant, and more. 

We are also interested in understanding the importance and limitations of learning-based algorithms for PCR. Learned features have been shown to be superior to hand-crafted ones is various fields. However, learning brings with it the issue of transferability. When we train FCGF features on a dataset that was collected in a specific location, \eg Boston, an important question is whether they will also be usable in other locations, \eg Singapore? How much of the learning is in fact memorization of characteristics that are specific to a location? 

Another question is with respect to the benefit of using deep-learning for robust parameter estimation. Some recent deep learning based algorithms, namely DGR and PointDSC, have shown a significant improvement for calculating the registration parameters when used with \emph{weak} features. Yet, when used with strong features that are in themselves learned (\eg FCGF), it is less clear whether an additional benefit is gained from these learning techniques over simpler approaches. For example, classical RANSAC has been shown to achieve comparable accuracy to the novel algorithms. Yet, prior works \cite{DGR,PointDSC,TEASER} claim that this requires significantly higher running time. Here, we suggest that this may not be a completely fair comparison: While basic RANSAC may be slow, many improvements have been suggested to it over the years for increasing its speed and accuracy. In this work, we evaluate combinations of several such improvements, and end up with a variant of RANSAC that is both more accurate \emph{and} faster than the current state-of-the-art robust estimation algorithms. We show that the accuracy of our presented framework can be further improved with an outlier filtering method, which we propose and name Grid-Prioritized Filtering (GPF).

\section{Related Work}
\label{sec:related}

Algorithms for rigid registration can roughly be divided into \emph{local} and \emph{global} ones. Local registration algorithms are based on the assumption that the motion is small. Global registration algorithms aim to handle any relative motion, but might be less accurate. Often their results are refined by running a local registration algorithm. 

\textbf{Local Registration:} Iterative Closest Points (ICP)~\cite{ICP} is one of the earliest successful approaches to local point cloud registration, and it remains popular to this day. The ICP algorithm has been developed in various different directions \cite{REVIEW}. Chen and Medioni~\cite{ICPChen} replaced the point-to-point loss function of ICP with a point-to-plane one, by using local normals. Segal \etal~\cite{GICP} presented the popular Generalized-ICP (G-ICP)~\cite{GICP} approach, which reformulated point-to-plane ICP in probabilistic terms and achieved improved accuracy. Rusinkiewicz recently suggested symmetric-ICP~\cite{SymICP}, which uses a surface-to-surface distance function that treats both point clouds symmetrically. It has been demonstrated to be superior to G-ICP in accuracy, and to have larger convergence basins.  Drory \etal~\cite{BBR} presented Best-Buddies Registration, specifically BBR-F, which uses a set of mutual-nearest-neighbors in the registration to improve accuracy.

\textbf{Global Registration:} A successful strategy for global registration is to generate a set of point-matches based on local descriptors, and estimate a motion from these matches. A popular classical descriptor is FPFH~\cite{FPFH} which uses histograms of gradients of neighboring points.

As in other fields, learned features have been shown to be superior to hand-crafted ones 
 \cite{PointNetLK,Sarode2019PCRNetPC,yew2020-RPMNet,Hertz20PointGMM,yuan2020deepgmr,DeepICP,DCP,PRNet}. Various such descriptor have been suggested, e.g. \cite{HRegNet, PREDATOR, FCGF}. Fully Convolutional Geometric Features (FCGF)~\cite{FCGF} are based on sparse convolutions over a voxelized representation of the point cloud. The FCGF network is very fast, and produces dense features.

\textbf{Robust optimization}: The set of descriptor matches typically includes a significant fraction of \emph{outliers}, which must be taken into consideration when estimating the relative motion. This can be done for example by using robust loss functions and algorithms, or by filtering the set of matches to remove outliers \cite{yang2017performance}. 
RANSAC~\cite{RANSAC} is a popular method, which works by repeatedly sampling a minimal set of point-matches, estimating a motion from the sample, and calculating its score by the fraction of matches that agree with this motion. This is repeated until a preset number of iterations is performed, or until early stopping occurs when the best-so-far motion has a fraction of inliers that is sufficient (relative to a confidence value supplied by the user~\cite{GC_RANSAC}). This simple framework has been greatly enhanced over the years, improving RANSAC in both speed and accuracy. 

PROSAC~\cite{PROSAC} performs a prioritized selection of candidate sets. It accepts the putative pairs sorted according to a quality measure, and orders the selection of sets so that sets with higher quality pairs are examined earlier. This simultaneously makes RANSAC faster and more accurate, by making it more likely that a good model is found early. 

LO-RANSAC~\cite{LO_RANSAC} adds a local-optimization step: when a best-so-far model is found, its inliers are used to find a better model, for example by performing RANSAC only on the inliers. Local optimization can be repeated several times, as long as the best-so-far model keeps improving significantly. Though the local-optimization step is expensive, it is only performed a few times over the run time of the RANSAC algorithm, and so its amortized time is small. The recently proposed GC-RANSAC~\cite{GC_RANSAC} uses a Markov-Random Field formulation and solves it with Graph-Cuts to divide pairs into inliers and outliers. 

Another important addition to RANSAC are early rejection methods, which can be applied quickly to reject a minimal set without going through the full scoring stage. We consider two such methods: Sequential Probability Ratio Test (SPRT)~\cite{SPRT}, a general domain method, and Edge-Length Consistency (ELC)~\cite{PointDSC}, which is specific to PCR. 

 In addition to producing local descriptors, deep-learning has also been used for robust estimation. Deep Global Registration (DGR)~\cite{DGR} is based on training a second FCGF-like deep network for the task of recognizing outliers. PointDSC~\cite{PointDSC} too is based on a second network, but not to simply recognize outliers. Instead, it learns an embedding space where one can locate groups of mutually-consistent pairs, that can be used to generate candidate motions. PointDSC integrated ELC into the neural network, to encourage spatial consistency. 
 
 A novel  approach that is not based on deep learning is TEASER, which is based on truncated least squares estimation and semi-definite relaxation. TEASER++ is a faster version that is based on Graduated Non-Convexity. 
 
\noindent {\bf Dataset Generation.}
Fontana \etal~\cite{Balanced} present a collection of datasets to be used as a benchmark for registration algorithms, and specify the method for the creation of these datasets. Unlike them, we focus specifically on LiDAR point cloud datasets, and registration sets that are challenging for global registration. We adopt their idea of achieving a balanced set of relative motions by random sampling. However, in their method a random motion is applied to an existing point cloud, thus creating a synthetic sample. Instead, we produce natural samples by selecting a pair of point clouds from a recorded sequence, so that their relative motion is as close to the randomly selected one as possible. 

Huang \etal~\cite{PREDATOR} present the 3DLoMatch set, that contains pairs of low-overlap scans from the 3DMatch~\cite{3DMatch} dataset. They define an overlap of between $10\%$ and $30\%$ as low. 
We set the minimum-overlap of our registration sets to $20\%$, which is in the same range. In line with their findings, our experiments show that low-overlap is a strong indicator for registration failure.

\section{Balanced LiDAR Registration Sets}
\label{sec:method_datasets}

Popular registration benchmarks for the automotive LiDAR setting have become too easy for the newest registration algorithms (see \cref{fig:kitti_saturation}). We believe the main cause for that are the simple heuristics used for selecting frame-pairs for registration: a constant offset in space or time, which is typically not very large (e.g. 10 meters, or 1 second).

\begin{wrapfigure}{r}{0.5\linewidth}
\begin{center}
\begin{tabular}{c}
\includegraphics[width=0.5\columnwidth]{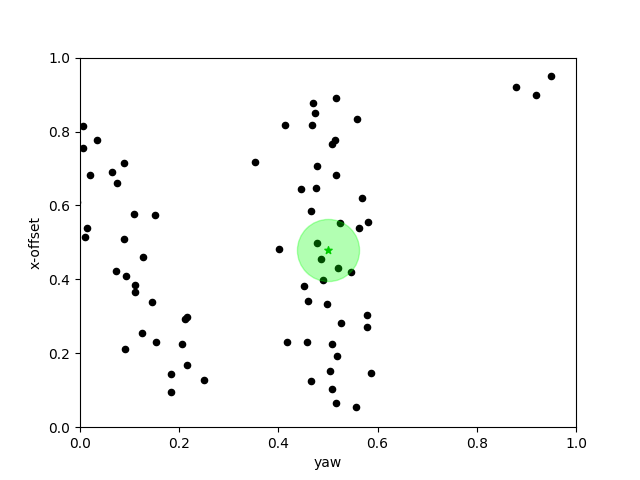}
\end{tabular}
\end{center}
\caption{\textbf{Selection of Balanced Registration Set.} Toy example of our selection method, using a 2DOF motion model (instead of 6DOF). Each black point represents the relative motion between a \emph{frame-pair}. The space of all motions is normalized into the unit square. Iteratively, we randomly sample a location (\emph{green asterisk}), and select one of the frame-pairs that is close enough to this location (within \emph{green circle}).}
\label{fig:selection_procedure}
\end{wrapfigure}

How could we instead select a more interesting set of frame-pairs? A naive approach would be to enumerate all possible frame-pairs in each driving-sequence, and then select randomly from them. This approach has two problems: first, many frame-pairs have no overlap, making registration impossible. Second, and more importantly, for a large majority of frame-pairs, the relative motion between them is simple, e.g. "small offset, no rotation". 

We suggest a different approach: sample uniformly from the space of motions. We think of the space of all relative motions as a 6-dimensional hyper-cube, whose axes are $x$-offset, $y$-offset, $z$-offset, roll, pitch and yaw. Different areas in this cube represent different \emph{types} of motions: small-offset with large yaw, large-offset with small yaw but large pitch, etc. By sampling uniformly at random from this hyper-cube, we end up with a set of frame-pairs that is challenging and contains representatives of all the types of motion that appear in the LiDAR dataset.



\textbf{Generating a pool of candidates.} In theory, every pair of point-clouds from the dataset could be considered as a candidate for the registration set. Yet, the total number of pairs is quadratic in the size of the dataset making this impractical. To generate a reasonably sized candidate pool, we take each $k$th frame in a sequence to be a source frame. For each source frame we find the set of frames whose overlap with it is above $min\_overlap$, and randomly choose the target frame from this set. 

\textbf{Random selection of samples.} We wish to select uniformly at random from the space of all relative motions that appear in the candidate pool. We iteratively repeat the following procedure (demonstrated in \cref{fig:selection_procedure}): First, we normalize each axis of the 6D hyper-cube separately to the range [0,1], to overcome different ranges for different axes (x-offset, yaw, etc.). Then, we randomly generate a location in the unit hyper-cube. If our location is farther than a radius $r$ from any candidate, we discard it and generate another. Otherwise, we consider the set of candidates within a radius $r$. They represent essentially the same type of motion, and we choose between them according to a second criterion: which driving sequence they come from. This allows us to encourage a fair representation for each driving sequence in the dataset, which is important since different sequences often include  different challenges: highways vs. residential areas, daytime vs. nighttime etc. 

We find it important to discard random locations that are farther than $r$ from any candidate. Allowing such locations to select the candidate nearest to them would have distorted the distribution of samples that we select. For instance, candidates that lie next to a large empty region of the hyper-cube would have a much higher probability of being selected. 


\textbf{Balanced registration sets.} Various Automotive LiDAR datasets are available, including KITTI-Odometry~\cite{KITTI}, NuScenes~\cite{NuScenes}, Apollo-Southbay~\cite{Apollo} and others. We use our algorithm to create three registration sets, that we use in our experiments. The sets are built over the Apollo-Southbay and NuScenes datasets. We divide NuScenes into two parts: Boston and Singapore. We name our registration sets \emph{Apollo-Southbay-Balanced}, \emph{NuScenes-Boston-Balanced} and \emph{NuScenes-Singapore-Balanced}. We set $min\_overlap\!=\!0.2$ and $r\!=\!0.1$. Our sets are considerably larger than KITTI-10m (see~\cref{app:exp} for size table). We believe this is beneficial in training, and also allows finer-grain comparison between algorithms in testing. 

In \cref{fig:set_stats} we compare the distribution of samples in \emph{Apollo-Southbay-Balanced} to that in KITTI-10m. We show marginal distributions according to different parameters: time-offset, distance, overlap, roll, yaw and pitch. In all parameters, our set includes a wider range of values than KITTI-10m. This is especially evident for distance, which for KITTI-10m is by definition always approximately 10 meters, and in our set is a wide range, upto over 50 meters. KITTI-10m includes only high-overlap pairs, while our dataset contains a range, actually focusing on the harder, low-overlap cases. Regarding yaw, KITTI-10m includes only small rotations, while our dataset includes a wide range, up to 90 degree turns and even some complete U-turns. Our dataset also contains more samples with significant roll and pitch than KITTI-10m does.


\section{Grid-Prioritized Filtering (GPF)}
\label{sec:GPF}

Pre-filtering the set of putative pair-matches, to reduce the fraction of outliers in it, is an important step in various PCR algorithms, including RANSAC and TEASER++. Various heuristic methods appear in the literature, including mutual-nearest neighbors (MNN, a.k.a reciprocity check), and ratio test~\cite{SIFT}, which can be based on distances in feature space or x-y-z space. In our work we consider registration with relatively low overlap, and we find that in this setting special care needs to be taken to make sure that after filtering, we are still left with points that are well spread spatially. To encourage this, we divide the point cloud into a grid (in the x-y dimension), and perform filtering separately in each grid cell. Special care is taken to balance the number of points remaining in each cell. We take sort the pairs in each cell according to two criteria: is/isn't MNN, and $1^{st}$ to $2^{nd}$ neighbor distance ratio. We look at distances in feature space, which makes more extensive usage of the deep-learned features than just ascertaining nearest-neighbor pairs. We call our filtering method Grid-Prioritized Filtering (GPF), and present its full details in~\cref{sec:GPF_detailed}.

\section{Experiments}
\label{sec:experiments}

\begin{figure}
\begin{minipage}[t]{.45\textwidth}
  \centering
  \includegraphics[width=1\linewidth,valign=t]{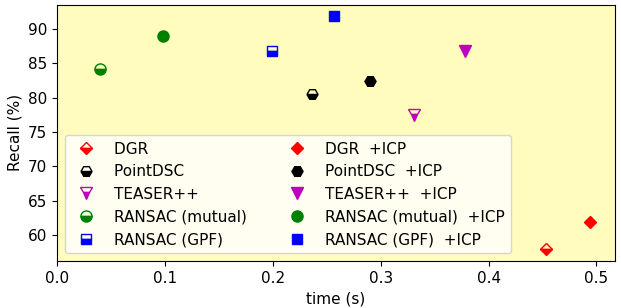}
  \captionof{figure}{\textbf{Comparison of registration algorithms on a balanced LiDAR dataset}. We use \emph{NuScenes-Boston-Balanced} to compare recent point-cloud registration algorithms. All algorithms use FCGF local-descriptors that were trained on this dataset. We show wall-time and recall, with and without ICP refinement. Advanced RANSAC is simultaneously \emph{faster and more accurate} than all other algorithms. Its two versions differ in the pre-filtering method used; The faster one \emph{(mutual)} uses mutual-nearest neighbors, and the more accurate one \emph{(GPF)} uses our proposed Grid-Prioritized Filtering.}
\label{fig:benchmark_B_to_B}
\end{minipage}
\begin{minipage}{.1\textwidth}
  \centering
\end{minipage}
\begin{minipage}{.1\textwidth}
  \centering
\end{minipage}
\begin{minipage}{.1\textwidth}
  \centering
\end{minipage}
\begin{minipage}{.1\textwidth}
  \centering
\end{minipage}
\begin{minipage}{.1\textwidth}
  \centering
\end{minipage}
\begin{minipage}[t]{.45\textwidth}
  \centering
  \includegraphics[width=1\linewidth,height=90pt,valign=t]{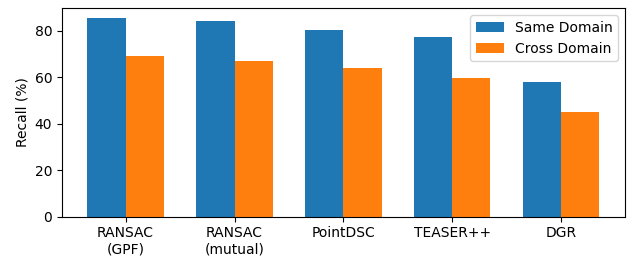}
  \captionof{figure}{\textbf{The effects of cross-domain testing}. When FCGF features are trained using a training set which is substantially different from the test set, we see a drop in accuracy. Here, the test set is from \emph{NuScenes-Boston-Balanced}, and the training set is either from \emph{NuScenes-Boston-Balanced} (same-domain, blue), or instead from \emph{Apollo-Southbay-Balanced} (cross-domain, orange). We see a drop in accuracy across all algorithms, of approximately 16 percentage points on average.}
\label{fig:cross_domain}  
\end{minipage}
\end{figure}

In this section we present several experiments, comparing different registration algorithms on the proposed LiDAR registration sets. All methods use FCGF~\cite{FCGF} deep-features trained on these sets, and differ in the robust estimation step. In some experiments the train-set and the test-set come from the same LiDAR dataset (same-domain), and in others from different datasets (cross-domain). This allows us to analyze the effect of cross-domain testing on deep feature accuracy. We compare the following algorithms: \textbf{Learned:} DGR~\cite{DGR}, PointDSC~\cite{PointDSC}, \textbf{algorithmic:} TEASER++~\cite{TEASER} and RANSAC. We tried various flavors of RANSAC (see~\cref{app:ablation}), and the best combination found includes:
\begin{enumerate}
    \item Prioritized selection of candidates (PROSAC), using the same priority order used in GPF
    \item Fast-rejection by edge-length consistency (ELC)
    \item Local-optimization step (LO-RANSAC), without graph-cuts
\end{enumerate}
 We compare two kinds of pre-filtering for RANSAC: mutual-nearest neighbors (MNN), and the novel GPF, with a $10\!\times\!10$ grid. TEASER++ also requires filtering, as it tends to get stuck indefinitely when receiving too many putative pair-matches as input. We use MNN for TEASER++ in all experiments, and add a second filtering with GPF when testing on \emph{Apollo-Southbay-Balanced} (see ahead). 
 
For each registration task we measure the rotation error (RE) and translation error (TE), defined as
\begin{equation}
\setlength\abovedisplayskip{0.1cm}
	\text{RE}(\mathbf{\hat R}) = \arccos \frac{\text{Tr}(\mathbf{\hat R^T}\mathbf{R^*})-1}{2},~~~
	\label{eq:re}
\end{equation}
\begin{equation}
	\text{TE}(\mathbf{\hat t}) = \left\Vert \mathbf{\hat t - t^*}\right\Vert_2.
    \label{eq:te}
\end{equation}
where $R^*,t^*$ is the ground-truth transformation.
We follow \cite{PointDSC} in defining a successful registration as one with RE$<$5 degrees and TE$<$0.6 meters (\ie, twice the voxel-grid spacing, see ahead). As a measure for the accuracy of an algorithm we use \emph{Recall}, which is the percentage of test samples for which registration succeeded. We report results before and after refinement, which is done with ICP (see~\cref{sec:local_reg} for experiments with some other refinement algorithms). Further implementation details can be found in~\cref{app:exp} . 

\subsection{Stress-Testing LiDAR registration}

In \cref{fig:benchmark_B_to_B}\footnote{also see table in~\cref{app:exp}} we present the results of using the \emph{NuScenes-Boston-Balanced} dataset to compare between DGR, PointDSC, TEASER++, and RANSAC with two pre-filtering algorithms: MNN (with max-iterations=1M, confidence=0.9995), and GPF(3.0) (with max-iterations=1M, and confidence=0.999). The fastest results are achieved by RANSAC with MNN filtering. The highest accuracy is achieved by RANSAC with GPF.

\begin{figure*}
\begin{center}
\includegraphics[width=1\textwidth]{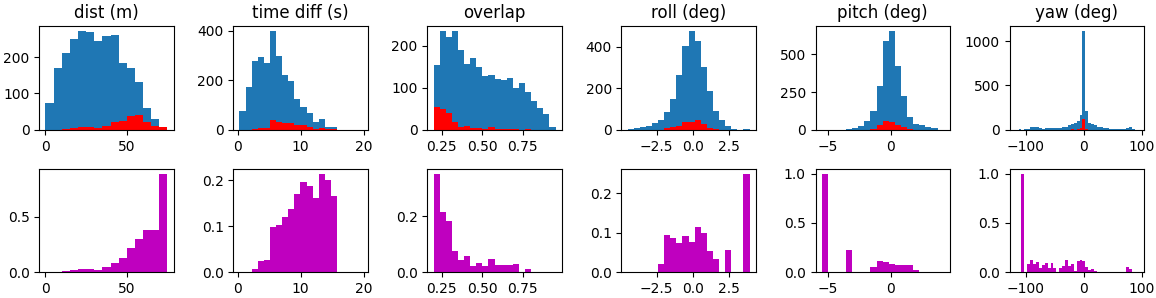}
\end{center}
   \caption{\textbf{Analysis of Failures.} We show the distribution of failed samples when running RANSAC (GPF) on the \emph{NuScenes-Boston-Balanced} dataset, with ICP refinement (see \cref{fig:benchmark_B_to_B}). On the top row we show the distribution of successful registrations (blue) and failed ones (red), according to several parameters. On the bottom row, we show the ratio of failures for each bin in the corresponding top row histogram. Large distance and small overlap emerge as the most influential parameters for failure. Other parameters seem to have little influence, except in the most extreme cases. }
\label{fig:failure_analysis}
\end{figure*}

By analyzing the failures of a PCR algorithm we can achieve an insight into its limitations. In~\cref{fig:failure_analysis} we show the distribution of failures in the previous experiemnt (with RANSAC+GPF). Distribution is according to several measures: distance between the point clouds, overlap, time offset, and three axes of rotation. One interesting conclusion is that the algorithms are extremely rotation invariant - the extent of rotation in the initial motion does not seem to be correlated with failure rates, except perhaps in a very few, very extreme cases. What emerge as the most influential parameters for failure are large distance and small overlap. Below an overlap of 0.35 we start to see a measurable increase in the failure rate. However, even with an overlap rate as low as 0.2, most of the registration attempts still succeed.

\begin{table}
\caption{Cross-Domain Evaluation}
\label{table:tight_results_train_A_test_B}
\begin{center}
\begin{tabular}{c  cccc}
\toprule
& \multicolumn{2}{c}{Algo. only} & \multicolumn{2}{c}{with ICP} \\
\cmidrule(r){2-5}
 & Recall & Time(s) & Recall & Time(s)
\\
\midrule\midrule
DGR & 44.95\% & 0.418 & 48.07\% & 0.462\\\midrule
PointDSC & 63.97\% & 0.234 & 66.78\% & 0.293\\\midrule
TEASER++ & 59.88\% & 0.146 & 71.99\%  & 0.213 \\\midrule
RANSAC (mutual)  & \uu{66.94\%} & \textbf{0.107} & \uu{74.31\%} & \textbf{0.171} \\\midrule
RANSAC (GPF)  & \textbf{69.14\%} & \uu{0.113} & \textbf{77.70\%} & \uu{0.177} \\\midrule
\bottomrule
\end{tabular}
\end{center}
\end{table}

In \cref{table:tight_results_train_A_test_B} we look at the setting of cross-domain testing. Here, all networks (FCGF, DGR and PointDSC) are trained on the \emph{Apollo-Southbay-Balanced} dataset, but the testing is on \emph{NuScenes-Boston-Balanced}. The ordering between algorithms remains the same as in the previous experiment, except here TEASER++ is faster than PointDSC. However, all recall values suffer a significant drop in the cross-domain case. \Cref{fig:cross_domain} visualizes this drop in accuracy for the case with ICP, showing a mean drop in recall of 16 percentage points. Cross-domain accuracies are significantly lower than the same-domain accuracies that we have seen in \cref{fig:benchmark_B_to_B}. We believe this shows that though FCGF features are quite transferable, some of their learning is location specific. In this experiment, we use GPF(2.0) with max-iterations=50K, and confidence=0.999. To allow clearer comparison, we use the same parameters also for the same-domain experiment in \cref{fig:cross_domain}.

\begin{table}
\caption{All Registration Sets Cross-Domain}
\label{tab:3_by_3}
\centering
\begin{tabular}{cccccc}\toprule
Test & Train & RANSAC & RANSAC &  PointDSC & TEASER++   \\
 &  & (GPF) & (mutual) &   &     \\\midrule\midrule
Apollo & Apollo & \textbf{98.97} & \uu{96.97} & 94.02 & 96.65 \\\midrule
Apollo & Boston & \textbf{93.84} & \uu{93.25} & 88.53 & 92.62  \\\midrule
Apollo & Singapore & \textbf{97.52} & 94.86 & 93.54 & \uu{95.16}  \\\midrule\midrule
Boston & Apollo & \textbf{77.70} & \uu{75.31} & 66.40 & 72.11  \\\midrule
Boston & Boston & \textbf{91.13} & \uu{89.39} &  82.37 & 86.88  \\\midrule
Boston & Singapore & \textbf{85.61} & \uu{80.79} & 75.39 & 79.63  \\\midrule\midrule
Singapore & Apollo & \textbf{88.43} & \uu{87.92} & 79.01 & 86.69 \\\midrule
Singapore & Boston & \textbf{91.47} & \uu{90.59} & 82.02 & 89.16  \\\midrule
Singapore & Singapore & \textbf{94.60} & \uu{93.75} & 90.59 & 93.29 \\\midrule
\bottomrule
\end{tabular}
\end{table}

\Cref{tab:3_by_3} presents a thorough test of our new datasets \emph{Apollo-Southbay-Balanced} (Apollo), \emph{NuScenes-Boston-Balanced} (Boston) and \emph{NuScenes-Singapore-Balanced} (Singapore). In each of the 9 experiments, one dataset is used for training, and another for testing. We test four algorithms: PointDSC, TEASER++, RANSAC with MNN filtering, and RANSAC with GPF. ICP is used for refinement in all cases. The highest result in each \emph{row} is in bold, the second underlined. 
Point clouds from the Apollo-Southbay dataset are approximately twice as large as those from NuScenes, and this ratio is maintained even after mutual-nearest neighbor filtering. As a result, TEASER++ tends to get stuck often ($\sim\!\!\!\!15\%$ of cases) when working on Apollo-Southbay point clouds. To overcome this, we use two mechanisms. First, a stricter filtering than usual for TEASER++: we first filter with MNN, and then filter with GPF, keeping a maximum of 2000 pairs. Second, we use a time-out of 10 seconds, after which registration is marked as failed. This happens very rarely (less than 0.1\% of cases). The larger point-clouds in Apollo-Southbay also affect our settings for RANSAC+GPF. We use GPF(1.0) when testing on Apollo-Southbay, and GPF(2.0) when testing on NuScenes. All other settings for RANSAC are as in the previous experiment.

In all cases, we see that accuracy drops when testing cross-domain. In addition, we can see that \emph{Apollo-Southbay-Balanced} is in a sense the simplest: it achieves the highest same-domain and cross-domain test results, but when networks are trained on it, they achieve the lowest cross-domain accuracy. Training on the \emph{NuScenes-Singapore-Balanced} dataset, on the other hand, leads to the highest cross domain accuracies. As far as algorithm comparison, RANSAC (GPF) is the most accurate, and RANSAC (mutual) the second except in one setting where TEASER++ is the second. Both RANSAC variants are also faster than the other algorithms in all cases (see~\cref{app:exp} for running times).

\section{Conclusion}
\label{sec:conclusion}

We thoroughly test novel PCR algorithms on challenging benchmarks in the automotive LiDAR setting. We find that deep-learned features such as FCGF\footnote{Though our work focuses specifically on FCGF, our methods are also applicable to other types of learned features \cite{D3Feat,PREDATOR}} suffer from partial reduction in accuracy when tested in a locale different from the one where they were trained.

We compare various robust parameter estimation algorithms, and find that the recent deep-learning based ones in fact do not achieve the best results. Instead, we find a variant of RANSAC which is both faster and more accurate than all other competitors. We also suggest an outlier filtering method, GPF, that further improves its accuracy. 
To improve current benchmarks, we have also introduced an algorithm for the selection of challenging frame-pairs from automotive LiDAR datasets. We believe it will be useful for future research, both for benchmarking PCR algorithms, and for training learned ones.

\medskip

{\small
\bibliographystyle{ieee_fullname}
\bibliography{egbib}

\begin{thebibliography}{10}\itemsep=-1pt

\bibitem{PointNetLK}
Yasuhiro Aoki, Hunter Goforth, Rangaprasad~Arun Srivatsan, and Simon Lucey.
\newblock Pointnetlk: Robust \& efficient point cloud registration using
  pointnet.
\newblock In {\em The IEEE Conference on Computer Vision and Pattern
  Recognition (CVPR)}, June 2019.

\bibitem{PointDSC}
Xuyang Bai, Zixin Luo, Lei Zhou, Hongkai Chen, Lei Li, Zeyu Hu, Hongbo Fu, and
  Chiew-Lan Tai.
\newblock Pointdsc: Robust point cloud registration using deep spatial
  consistency.
\newblock In {\em Proceedings of the IEEE/CVF Conference on Computer Vision and
  Pattern Recognition (CVPR)}, pages 15859--15869, June 2021.

\bibitem{D3Feat}
Xuyang Bai, Zixin Luo, Lei Zhou, Hongbo Fu, Long Quan, and Chiew-Lan Tai.
\newblock D3feat: Joint learning of dense detection and description of 3d local
  features.
\newblock In {\em Proceedings of the IEEE/CVF Conference on Computer Vision and
  Pattern Recognition (CVPR)}, June 2020.

\bibitem{GC_RANSAC}
Daniel Barath and Jiří Matas.
\newblock Graph-cut ransac.
\newblock In {\em Proceedings of the IEEE Conference on Computer Vision and
  Pattern Recognition (CVPR)}, June 2018.

\bibitem{ICP}
Paul~J. Besl and Neil~D. McKay.
\newblock A method for registration of 3-d shapes.
\newblock {\em IEEE Trans. Pattern Anal. Mach. Intell.}, 14(2):239--256, Feb.
  1992.

\bibitem{NuScenes}
Holger Caesar, Varun Bankiti, Alex~H. Lang, Sourabh Vora, Venice~Erin Liong,
  Qiang Xu, Anush Krishnan, Yu Pan, Giancarlo Baldan, and Oscar Beijbom.
\newblock nuscenes: A multimodal dataset for autonomous driving.
\newblock {\em arXiv preprint arXiv:1903.11027}, 2019.

\bibitem{ICPChen}
Yang Chen and G{\'e}rard Medioni.
\newblock Object modelling by registration of multiple range images.
\newblock {\em Image Vision Comput.}, 10(3):145--155, Apr. 1992.

\bibitem{DGR}
Christopher Choy, Wei Dong, and Vladlen Koltun.
\newblock Deep global registration.
\newblock In {\em CVPR}, 2020.

\bibitem{minkowski}
Christopher Choy, JunYoung Gwak, and Silvio Savarese.
\newblock 4d spatio-temporal convnets: Minkowski convolutional neural networks.
\newblock In {\em Proceedings of the IEEE Conference on Computer Vision and
  Pattern Recognition}, pages 3075--3084, 2019.

\bibitem{FCGF}
Christopher Choy, Jaesik Park, and Vladlen Koltun.
\newblock Fully convolutional geometric features.
\newblock In {\em ICCV}, 2019.

\bibitem{PROSAC}
O. Chum and J. Matas.
\newblock Matching with prosac - progressive sample consensus.
\newblock In {\em 2013 IEEE Conference on Computer Vision and Pattern
  Recognition}, volume~2, pages 220--226, jun 2005.

\bibitem{LO_RANSAC}
Ondrej Chum, Jiri Matas, and Josef Kittler.
\newblock Locally optimized ransac.
\newblock In {\em DAGM-Symposium}, volume 2781 of {\em Lecture Notes in
  Computer Science}, pages 236--243. Springer, 2003.

\bibitem{BBR}
Amnon Drory, Tal Shomer, Shai Avidan, and Raja Giryes.
\newblock Best buddies registration for point clouds.
\newblock In {\em Proceedings of the Asian Conference on Computer Vision
  (ACCV)}, November 2020.

\bibitem{RANSAC}
Martin~A. {Fischler} and Robert~C. {Bolles}.
\newblock Random sample consensus: a paradigm for model fitting with
  applications to image analysis and automated cartography.
\newblock {\em Communications of The ACM}, 1981.

\bibitem{Balanced}
Simone Fontana, Daniele Cattaneo, Augusto~L. Ballardini, Matteo Vaghi, and
  Domenico~G. Sorrenti.
\newblock A benchmark for point clouds registration algorithms.
\newblock {\em Robotics and Autonomous Systems}, 140:103734, 2021.

\bibitem{KITTI}
Andreas Geiger, Philip Lenz, and Raquel Urtasun.
\newblock Are we ready for autonomous driving? the kitti vision benchmark
  suite.
\newblock In {\em Conference on Computer Vision and Pattern Recognition
  (CVPR)}, 2012.

\bibitem{Hertz20PointGMM}
A. {Hertz}, R. {Hanocka}, R. {Giryes}, and D. {Cohen-Or}.
\newblock Pointgmm: A neural gmm network for point clouds.
\newblock In {\em IEEE/CVF Conference on Computer Vision and Pattern
  Recognition (CVPR)}, pages 12051--12060, 2020.

\bibitem{PREDATOR}
Shengyu Huang, Zan Gojcic, Mikhail Usvyatsov, Andreas Wieser, and Konrad
  Schindler.
\newblock Predator: Registration of 3d point clouds with low overlap.
\newblock In {\em Proceedings of the IEEE/CVF Conference on Computer Vision and
  Pattern Recognition (CVPR)}, pages 4267--4276, June 2021.

\bibitem{SIFT}
David~G. Lowe.
\newblock Distinctive image features from scale-invariant keypoints.
\newblock {\em Int. J. Comput. Vision}, 60(2):91--110, Nov. 2004.

\bibitem{HRegNet}
Fan Lu, Guang Chen, Yinlong Liu, Lijun Zhang, Sanqing Qu, Shu Liu, and Rongqi
  Gu.
\newblock Hregnet: A hierarchical network for large-scale outdoor lidar point
  cloud registration.
\newblock In {\em Proceedings of the IEEE/CVF International Conference on
  Computer Vision (ICCV)}, pages 16014--16023, October 2021.

\bibitem{DeepICP}
Weixin Lu, Guowei Wan, Yao Zhou, Xiangyu Fu, Pengfei Yuan, and Shiyu Song.
\newblock Deepvcp: An end-to-end deep neural network for point cloud
  registration.
\newblock {\em 2019 IEEE/CVF International Conference on Computer Vision
  (ICCV)}, Oct 2019.

\bibitem{Apollo}
Weixin Lu, Yao Zhou, Guowei Wan, Shenhua Hou, and Shiyu Song.
\newblock L3-net: Towards learning based lidar localization for autonomous
  driving.
\newblock In {\em Proceedings of the IEEE Conference on Computer Vision and
  Pattern Recognition}, pages 6389--6398, 2019.

\bibitem{SPRT}
Jiri Matas and Ondrej Chum.
\newblock Randomized {RANSAC} with sequential probability ratio test.
\newblock In {\em 10th {IEEE} International Conference on Computer Vision
  {(ICCV} 2005), 17-20 October 2005, Beijing, China}, pages 1727--1732. {IEEE}
  Computer Society, 2005.

\bibitem{REVIEW}
Fran{\c{c}}ois Pomerleau, Francis Colas, and Roland Siegwart.
\newblock A review of point cloud registration algorithms for mobile robotics.
\newblock {\em now}, 2015.

\bibitem{SymICP}
Szymon Rusinkiewicz.
\newblock A symmetric objective function for {ICP}.
\newblock {\em ACM Transactions on Graphics (Proc. SIGGRAPH)}, 38(4), July
  2019.

\bibitem{FPFH}
Radu~Bogdan Rusu, Nico Blodow, and Michael Beetz.
\newblock Fast point feature histograms (fpfh) for 3d registration.
\newblock In {\em ICRA}, 2009.

\bibitem{Sarode2019PCRNetPC}
Vinit Sarode, Xueqian Li, Hunter Goforth, Yasuhiro Aoki, Rangaprasad~Arun
  Srivatsan, Simon Lucey, and Howie Choset.
\newblock Pcrnet: Point cloud registration network using pointnet encoding.
\newblock {\em ArXiv}, abs/1908.07906, 2019.

\bibitem{GICP}
Aleksandr Segal, Dirk Hähnel, and Sebastian Thrun.
\newblock Generalized-icp.
\newblock In Jeff Trinkle, Yoky Matsuoka, and José~A. Castellanos, editors,
  {\em Robotics: Science and Systems}. The MIT Press, 2009.

\bibitem{DCP}
Yue Wang and Justin~M. Solomon.
\newblock Deep closest point: Learning representations for point cloud
  registration.
\newblock In {\em The IEEE International Conference on Computer Vision (ICCV)},
  October 2019.

\bibitem{PRNet}
Yue Wang and Justin~M. Solomon.
\newblock Prnet: Self-supervised learning for partial-to-partial registration.
\newblock In Hanna~M. Wallach, Hugo Larochelle, Alina Beygelzimer, Florence
  d'Alch{\'{e}}{-}Buc, Emily~B. Fox, and Roman Garnett, editors, {\em Advances
  in Neural Information Processing Systems 32: Annual Conference on Neural
  Information Processing Systems 2019, NeurIPS 2019, 8-14 December 2019,
  Vancouver, BC, Canada}, pages 8812--8824, 2019.

\bibitem{TEASER}
Heng Yang, Jingnan Shi, and Luca Carlone.
\newblock Teaser: Fast and certifiable point cloud registration.
\newblock {\em IEEE Transactions on Robotics}, 37(2):314--333, 2021.

\bibitem{yang2017performance}
Jiaqi Yang, Ke Xian, Yang Xiao, and Zhiguo Cao.
\newblock Performance evaluation of 3d correspondence grouping algorithms.
\newblock In {\em 2017 International Conference on 3D Vision (3DV)}, pages
  467--476, 2017.

\bibitem{yew2020-RPMNet}
Zi~Jian Yew and Gim~Hee Lee.
\newblock Rpm-net: Robust point matching using learned features.
\newblock In {\em Conference on Computer Vision and Pattern Recognition
  (CVPR)}, 2020.

\bibitem{yuan2020deepgmr}
Wentao Yuan, Benjamin Eckart, Kihwan Kim, Varun Jampani, Dieter Fox, and Jan
  Kautz.
\newblock Deepgmr: Learning latent gaussian mixture models for registration.
\newblock In {\em ECCV}, 2020.

\bibitem{3DMatch}
Andy Zeng, Shuran Song, Matthias Nie{\ss}ner, Matthew Fisher, Jianxiong Xiao,
  and Thomas Funkhouser.
\newblock 3dmatch: Learning local geometric descriptors from rgb-d
  reconstructions.
\newblock In {\em CVPR}, 2017.

\bibitem{Open3D}
Qian-Yi Zhou, Jaesik Park, and Vladlen Koltun.
\newblock {Open3D}: {A} modern library for {3D} data processing.
\newblock {\em arXiv:1801.09847}, 2018.

\end{thebibliography}
}



\newpage
\begin{appendices}
\crefalias{section}{appendix}
\counterwithin{figure}{section}
\counterwithin{table}{section}
\section{Additional Experiment Details}
\label{app:exp}

The number of samples in each set is shown in \Cref{table:dataset_sizes}.

\begin{table}
\caption{Balanced registration set sizes}
\label{table:dataset_sizes}
\begin{center}
\begin{tabular}{c*{3}{c}}\toprule
 & Train & Validation & Test
\\\midrule\midrule
KITTI-10m & 1338 & 200 & 555 \\\midrule
NuScenes-Boston-Balanced & 4032 & 384 & 2592 \\\midrule
NuScenes-Singapore-Balanced & 4032 & 384 & 2592 \\\midrule
Apollo-Southbay-Balanced & 4032 & 288 & 7008 \\
\bottomrule
\end{tabular}
\end{center}

\end{table}

\Cref{table:tight_results_train_B_test_B} presents the results of the same-domain experiment in tabular form (corresponds to~\cref{fig:benchmark_B_to_B} in main article). The right side of the table shows results with ICP refinement, the left side without. The best result in each column is in bold and the second best is underlined.

\begin{table}
\caption{Evaluation of Registration Algorithms}
\label{table:tight_results_train_B_test_B}
\begin{center}
\begin{tabular}{ccccc}\toprule
& \multicolumn{2}{c}{Algo. only} & \multicolumn{2}{c}{with ICP} \\\cmidrule(r){2-3}\cmidrule(r){4-5}
 & Recall & Time(s) & Recall & Time(s)
\\\midrule\midrule
DGR & 57.91\% &  0.453 & 61.81\% &  0.494\\\midrule
PointDSC & 80.56\% & 0.236 & 82.48\% & 0.290 \\\midrule
TEASER++ & 77.43\% & 0.331 & 86.88\% & 0.378\\\midrule
RANSAC (mutual) & \uu{84.14\%} & \textbf{0.040} & \uu{89.01\%} & \textbf{0.099} \\\midrule
RANSAC (GPF) & \textbf{86.88\%} & \uu{0.199} & \textbf{91.90\%} & \uu{0.257} \\
\bottomrule
\end{tabular}
\end{center}
\end{table}

In \cref{tab:3_by_3_runtime} we show the running times (in seconds) for the All-Set Cross Domain experiment (corresponds to \cref{tab:3_by_3} in main paper). Fastest in each row (always RANSAC mutual) is in bold, second fastest (always RANSAC+GPF) is underlined.

\begin{table}
\caption{Running Times for All Registration Sets Cross-Domain Experiment}
\label{tab:3_by_3_runtime}
\centering
\begin{tabular}{cccccc}\toprule
Test & Train & RANSAC & RANSAC &  PointDSC & TEASER++   \\
 &  & (GPF) & (mutual) &   &     \\\midrule\midrule
Apollo & Apollo & \uu{0.326} & \textbf{0.292} & 0.691 & 0.781 \\\midrule
Apollo & Boston & \uu{0.336} & \textbf{0.330}  & 0.725 & 0.354  \\\midrule
Apollo & Singapore & \uu{0.346} & \textbf{0.317} & 0.702  & 0.449  \\\midrule\midrule
Boston & Apollo & \uu{0.177} & \textbf{0.171}  & 0.451 & 0.277 \\\midrule
Boston & Boston & \uu{0.157} & \textbf{0.098}  & 0.432 & 0.486  \\\midrule
Boston & Singapore  & \uu{0.177} & \textbf{0.124} & 0.437 & 0.477  \\\midrule\midrule
Singapore & Apollo & \uu{0.228} & \textbf{0.202} & 0.616 & 0.250  \\\midrule
Singapore & Boston  & \uu{0.224} & \textbf{0.147} & 0.608 & 0.258   \\\midrule
Singapore & Singapore  & \uu{0.237} & \textbf{0.119} & 0.589 & 0.846  \\\midrule
\bottomrule
\end{tabular}
\end{table}

We mention in the paper that \emph{Apollo-Southbay-Balanced} has larger point clouds than the other datasets we use, and that this is also true after mutual-nearest neighbor filtering. In~\cref{tab:inlier_count} we show the number of putative pair-matches for different experiments, before and after mutual-nearest neighbor (MNN) filtering. The datasets are \emph{Apollo-Southbay-Balanced} (Apollo), \emph{NuScenes-Boston-Balanced} (Boston) and \emph{NuScenes-Singapore-Balanced} (Singapore). The values shown are averaged over all samples in each dataset.

\begin{table}
\caption{Pair-Match Set Sizes}
\label{tab:inlier_count}
\begin{center}
\begin{tabular}{cccc}\toprule
Test & Train & Initial & MNN-filtered 
\\\midrule\midrule
Apollo & Apollo & 23520  & 2123   \\\midrule
Apollo & Boston & 23520  &  1717 \\\midrule
Apollo & Singapore & 23520  & 1830  \\\midrule\midrule
Boston & Apollo & 8091  & 766  \\\midrule
Boston & Boston & 8091   &   837   \\\midrule
Boston & Singapore & 8091  & 841  \\\midrule\midrule
Singapore & Apollo & 10335  & 1106  \\\midrule
Singapore & Boston & 10335  &  1104 \\\midrule
Singapore & Singapore & 10335 & 1198  \\\midrule
\bottomrule
\end{tabular}
\end{center}
\end{table}

\subsection{Implementation Details} 
Calculating FCGF features requires all points to lie on a grid. Thus, we start all registration algorithms by down-sampling with an 0.3 meter voxel-grid filter (following~\cite{FCGF, DGR, PointDSC}). We continue by calculating FCGF features and finding nearest-neighbors in the feature space. When reporting running time we omit the time taken by this pre-processing. 

\textbf{Code\footnote{See~\cref{app:code} for links and license information.}:} for RANSAC we use the GC-RANSAC~\cite{GC_RANSAC} code base, which is efficiently implemented and offers multiple options (PROSAC, local-optimization, etc.). We added an ELC implementation based on the one in open3d (version 0.13) \cite{Open3D}. We've also tried the open3d implementation of RANSAC (see appendix), which offers fewer options, and is somewhat slower, though still quite fast. We run the GC-RANSAC code with \emph{distance\_ratio=0.6} and \emph{spatial\_coherence\_weight=0}, which effectively makes it LO-RANSAC and not GC-RANSAC. We also enable PROSAC and ELC. We set outlier filtering parameters for each experiment separately, to demonstrate RANSAC's ability to achieve both the fastest and most accurate results. %
For ICP we use open3D, with \emph{threshold=0.6}. 

For DGR, PointDSC and TEASER++ we use the official implementations, with slight modifications. 
We use our own implementation for training FCGF features. 

The number of training epochs was selected according to preliminary tests (not shown), to a level where further improvement is very slow. The values are: 400 epochs for FCGF, 50 for PointDSC and 40 for DGR (whose training is considerably slower). We also changed the rotation augmentation scheme to make more sense in the automotive LiDAR setting: instead of general rotations in all axes, we augment with nearly-planar rotations, where yaw is in the range $\pm180$ degrees, but pitch and roll are only up to $\pm5$ degrees.

We use two machines for our experiments:
\begin{enumerate}[label=\Alph*]
    \item GPU: 4x Titan X, CPU: 20-core 2.20GHz Xeon E5-2630
    \item GPU: GTX 980 Ti, CPU: 8-core 4.00GHz i7-6700K 
\end{enumerate}
Most of our tests are performed on machine A, using a single GPU. TEASER++ code is run on machine B, due to its code failing to work on machine A. To compare running time, we extrapolate TEASER++'s presumptive running time on machine A. To do so, we calculate a normalizing ratio by running RANSAC on both machines. In the appendix we analyze the differences in CPU and GPU running times across machines.

\section{Detailed Description of Grid-Prioritized Filtering (GPF)}
\label{sec:GPF_detailed}

\begin{figure*}
\centerline{\includegraphics[width=1\textwidth]{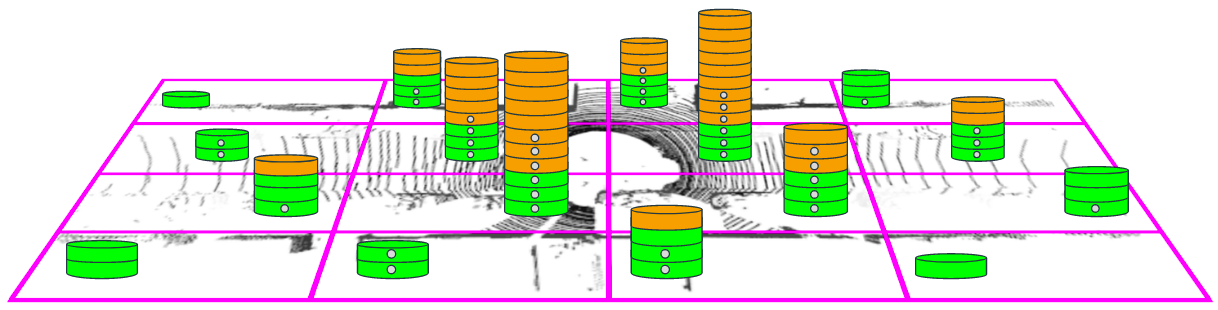}}
\caption{\textbf{Grid-Prioritized Filtering (GPF).} GPF is a filtering algorithm used to select a subset of putative point-matches that are both high-quality and maximally spatially spread. This is achieved by dividing the source point-cloud into a grid of cells on the x-y plane, and selecting approximately the same number of matches from each cell. In the diagram, each match is represented by a disk, with those on the bottom, colored green, representing the ones that were selected by GPF. Within each cell, matches are ordered bottom-to-top by their estimated quality, based on analysis of the feature-space distance between the pair. Mutual nearest-neighbors (disks marked with small white circles) are selected first. Then, non-mutual. The secondary criterion for prioritizing is \cref{eq:rat_dist} (ratio between distance to 1st and 2nd nearest neighbor).}
\label{fig:GPF}
\end{figure*}

We propose the Grid-Prioritized Filtering (GPF) method to explicitly ensure spatial spread in the selected pairs. As illustrated in \cref{fig:GPF}, GPF works by dividing the source point cloud into an $M \times M$ grid in the x-y plane. Then, $\ell$ matches are selected from each grid cell (or all matches if there are fewer than $\ell$ in the cell). The priority of pairs to select follows two criteria: First, matches that are MNNs are preferred. The secondary ordering criterion is the ratio $S$: 
\begin{align}
\label{eq:rat_dist}
    S(p) = \frac{d(p,q_2)}{d(p,q_1)},
\end{align}
where $P,Q$ are point clouds, $p\!\in\!P$, $q_1, q_2 \!\in\! Q$, $q_1$ is the nearest neighbor to $p$ in $Q$ , $q_2$ is the second-nearest, and $d()$ is the $L_2$ distance.

The number of pairs per cell, $\ell$ is determined by the total requested number, $R$. The simple calculation $\ell\!=\!R/M^2$ is only valid when all cells contain at-least $\ell$ pairs. Instead, we perform a quick binary search to find the value of $\ell$ that brings the overall selected number closest to $R$. 
$R$ can be specified explicitly, but we believe that matching it to the properties of each point-cloud is preferable. To do so, we define it by: 
\vspace*{-0.4cm}
\begin{align}
R=\phi\cdot|\mathcal{N}|,
\end{align}
where $\mathcal{N}$ is the set of \emph{mutual} nearest neighbors for each cloud, and $\phi$ is the user supplied \emph{GPF factor}.
We use notation like GPF(2.0) to refer to running GPF with $\phi\!=\!2.0$. 

\section{Local Registration (refinement)}
\label{sec:local_reg}
Balanced datasets can also be used to compare local registration algorithms, such as ICP. Such algorithms take an initial coarse motion estimation, and refine it to achieve a high accuracy alignment. To use them with our balanced registration sets, we supply a standard set of \emph{initial motions}, produced by performing RANSAC registration with FCGF features. These initial motions are generally close enough to the ground truth motion to allow local registration algorithms to succees. In \cref{tab:refinement} we show the results of using the \emph{Apollo-Southbay-Balanced} dataset, and comparing three local registration algorithms: ICP~\cite{ICP}, symmetric-ICP~\cite{SymICP}, and BBR-F~\cite{BBR}. We use the official implementations of symmetric-ICP and BBR-F, and the open3d implementation of ICP. The point clouds are downsampled with a voxel-grid filter with a voxel size of 0.3 meters, and we set ICP's threshold to 0.6 meters (as we do in all experiments, following~\cite{DGR}). We report Recall, as well as translation error (TE) and rotation error (RE). We report mean, median and 95\textsuperscript{th} percentile of TE and RE, and these statistics are taken over \emph{all} test samples. The results show that ICP is more accurate than BBR-F, and both are considerably more accurate than symmetric-ICP. This differs from previous experiments in ~\cite{BBR} that used a subset of KITTI. We believe the central factor is overlap between point-clouds: small overlap is common in our sets but not in KITTI. ICP explicitly filters point pairs whose distance is above a threshold, and BBR-F uses spatial mutual-nearest neighbors. These elements apparently gives them an edge over Symmteric-ICP in this setting.

\begin{table}
\caption{Refinement Experiment}
\label{tab:refinement}
\begin{center}
\begin{tabular}{cccccccc}\toprule
 Algorithm & Recall & \multicolumn{3}{c}{TE (cm)} & \multicolumn{3}{c}{RE (deg)} \\\cmidrule(r){1-1}\cmidrule(r){2-2}\cmidrule(r){3-5}\cmidrule(r){6-8}
 && mean & 50\% & 95\% & mean & 50\% & 95\% 
\\\midrule
ICP & 98.99\% & 80.65 & 11.76 & 30.29 & 0.37 & 0.13 & 0.33 \\\midrule
BBR-F & 96.33\% & 86.98 & 15.10 & 52.84 & 0.47 & 0.19 & 0.66
\\\midrule
sym-ICP & 67.74\% & 548.85 & 17.68 & 3544.49 & 2.31 & 0.22 & 10.66
\\\bottomrule
\end{tabular}
\end{center}
\end{table}

\section{Ablation Studies}
\label{app:ablation}
\subsection{RANSAC}

\begin{figure}
\centering
\includegraphics[width=1\textwidth]{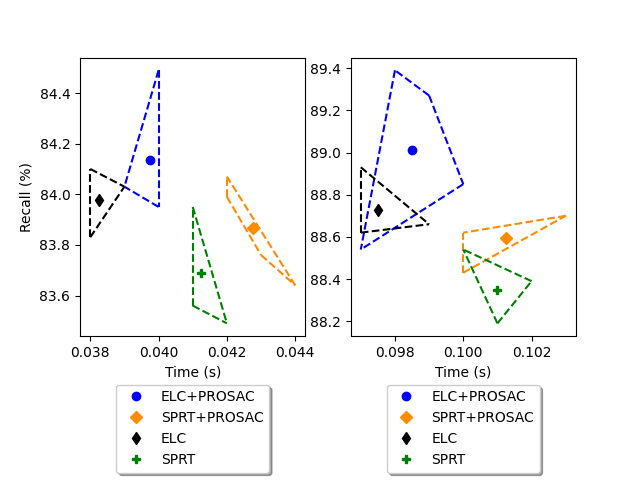}
 
\caption{\textbf{RANSAC Ablation: PROSAC and ELC/SPRT}. We show the accuracy and running time of different variants of RANSAC, with ICP (right) and without (left). For each setting, we repeat the run 4 times and show the spread of results by a polygon (the convex hull). We also show their mean. The best results are when we use both PROSAC and ELC.}
\label{fig:RANSAC_ablation_1}
\end{figure}

\begin{figure}
\centering
\includegraphics[width=1\textwidth]{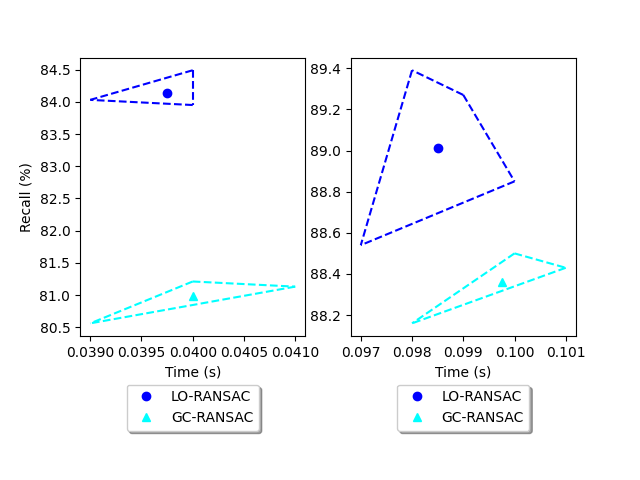}
 
\caption{\textbf{RANSAC Ablation: Local-Optimization}. Using the same visualization as \cref{fig:RANSAC_ablation_1}, we show LO-RANSAC is superior to GC-RANSAC in our setting. }
\label{fig:RANSAC_ablation_2}
\end{figure}

\begin{figure}
\centering
\includegraphics[width=1\textwidth]{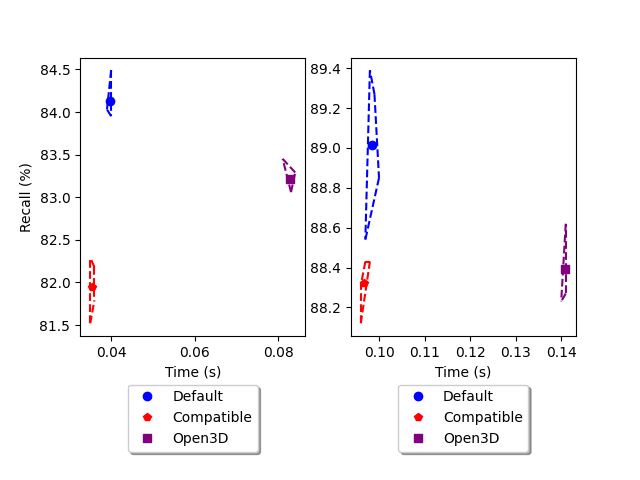}
 
\caption{\textbf{RANSAC code bases: Open-3D vs. GC-RANSAC}. We compare the open3d implementation of RANSAC (with ELC), to the GC-RANSAC implementation in two settings: "compatible" which is as similar as possible to open3d, and "default" which is what we use in most of our experiments. The open3d implementation is slower than the GC-RANSAC one. It also does not offer PROSAC and LO-RANSAC, causing it to be less accurate. However, it is still faster and more accurate than all other algorithm we tested (TEASER++, PointDSC, DGR).}
\label{fig:RANSAC_ablation_open3d}
\end{figure}

The version of RANSAC that we use in our experiments includes several improvements over classical RANSAC:
\begin{enumerate}
    \item Prioritized selection of candidate sets (PROSAC).
    \item Quick rejection of candidate sets (with ELC). 
    \item Local-Optimization step (LO-RANSAC).
\end{enumerate}
We perform ablation studies to show the importance of each element. We both train and test on \emph{NuScenes-Boston-Balanced}, and use the same settings as in the experiment shown in Tab. 2 of the main paper, for the nearest-neighbor filtering case. All variants of RANSAC tested in this section are both faster and more accurate than the other algorithms we consider in our paper: TEASER++, PointDSC and DGR.  The results of our first experiment are shown in \cref{fig:RANSAC_ablation_1}. We compare PROSAC to random selection of candidate sets, and in the quick rejection step, we compare ELC to SPRT. To show variance, we repeat each experiment 4 times, and plot both the mean and the convex hull of the 4 results. The results demonstrate that adding PROSAC improve accuracy but also adds to running time, and that replacing SPRT with ELC improves both accuracy and running time.

In \cref{fig:RANSAC_ablation_2} we show a comparison of LO-RANSAC to GC-RANSAC. In both cases we use PROSAC and ELC, and the only difference is the parameter \emph{spatial\_coherence\_weight}. To run LO-RANSAC we set it to 0. To run GC-RANSAC, we set it to its default value, 0.975. LO-RANSAC achieves higher recall than GC-RANSAC in our setting.  We also tested other values of the parameter (not shown), and the best accuracy was achieved with 0 (i.e. LO-RANSAC). 

In \cref{fig:RANSAC_ablation_open3d} we compare the open3d implementation of RANSAC to the GC-RANSAC implementation which we use for most experiments (we refer to it as \emph{GC-code}). The open3d implementation includes ELC, but does not include local-optimization and PROSAC. For the fairest comparison, we run the GC-code in a "compatible" setting, also using ELC but no local-optimization and no PROSAC. For reference, we also run the GC-code with our default setting (ELC, PROSAC and LO-RANSAC). Open3D is considerably slower than either GC-code setting. It is less accurate than our default setting of GC-code, but interestingly more accurate than the "compatible" setting. Possibly, this is due to differences in the implementation of early stopping.
Open3d RANSAC is both faster and more accurate than all other algorithms we tested, (compare \cref{fig:RANSAC_ablation_open3d} here to Fig.~6 in main paper).

\subsection{GPF}
\begin{figure}
\centering
\includegraphics[width=1\textwidth]{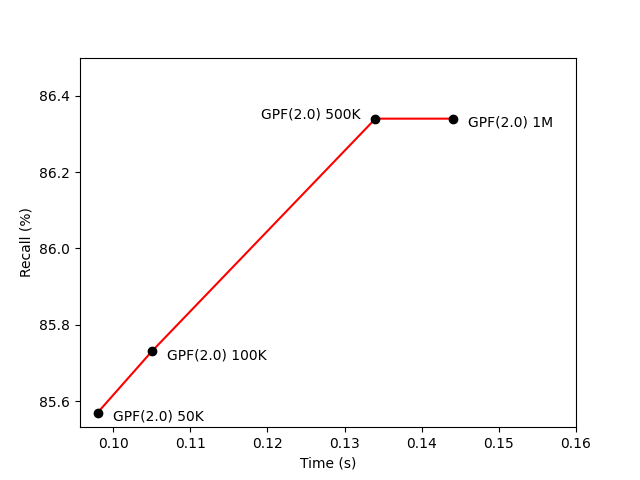}
\includegraphics[width=1\textwidth]{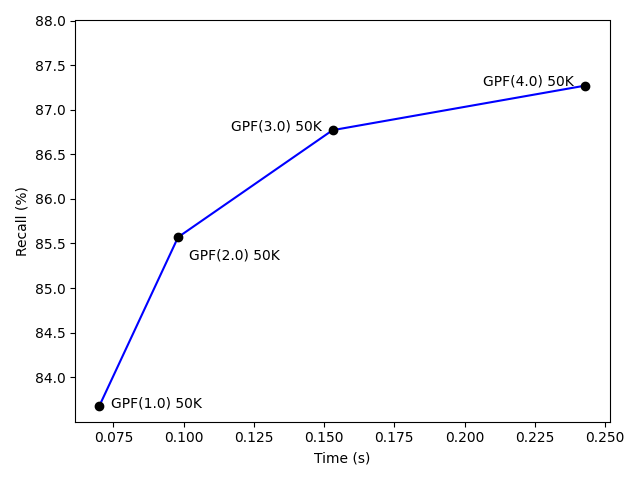}
 
\caption{\textbf{GPF Ablation}. We show the effects of different values of \emph{max-iteration} (top) and of $\phi$ (bottom). Increasing max-iterations improves accuracy only up to a point, after which accuracy plateaus while running time increases. Increasing $\phi$ improves accuracy and increases running time, and the plateau phenomenon is much less pronounced. }
\label{fig:GPF_ablation}
\end{figure}

In \cref{fig:GPF_ablation} we demonstrate the effect of the number of iterations and of the GPF parameter when running RANSAC+GPF. We can see that when adding iterations, running time always increases, but accuracy reaches saturation and plateaus at some point. Increasing the GPF parameter $\phi$, which corresponds to keeping a larger set of point-pairs, leads to an increase in both running time and in accuracy. However, the increase in accuracy does become considerably slower as we advance the parameter above $3.0$. In our main experiments we used the parameter values of $1.0$, $2.0$ and $3.0$.

\section{GPU and CPU Running Time on Different Machines.}
\label{sec:running_time}
Some of the registration algorithms that we compare rely mostly on GPU for processing (PointDSC, DGR), while others mostly use the CPU (TEASER++, RANSAC). Therefore, a comparison of running times between these algorithms depends on the specific machine being used. We demonstrate this in \cref{tab:GPU_CPU_times}, by running the same experiment on two machines. The machines that we use are:
\begin{enumerate}[label=\Alph*]
    \item GPU: 4x Titan X, CPU: 20-core 2.20GHz Xeon E5-2630
    \item GPU: GTX 980 Ti, CPU: 8-core 4.00GHz i7-6700K 
\end{enumerate}
On either machine, we use only one GPU for testing. The experiment consists of testing PointDSC and RANSAC on the \emph{NuScenes-Boston-Balanced} dataset (training was also performed on the same dataset). We report the running times on both machines. The ratio between the running times of PointDSC and RANSAC is different between the machines, reflecting the different mixes of CPU and GPU capabilities in each machine.  For this experiment, we used the open3D implementation of RANSAC.

\begin{table}
\caption{Running Times Comparison on Two Machines}
\label{tab:GPU_CPU_times}
\begin{center}
\begin{tabular}{cccc}\toprule
 Algorithm & Main & Machine A & Machine B  \\
 & Resource & Time (s) & Time (s)
 \\\midrule\midrule
PointDSC & GPU & 0.236 & 0.330  \\\midrule
RANSAC & CPU & 0.109 & 0.135  \\\midrule
Ratio PointDSC/RANSAC & & 2.44 & 2.17 \\
\bottomrule
\end{tabular}
\end{center}
\end{table}

\section{Code Bases}
\label{app:code}
In our work we make use the following code bases:
\textbf{FCGF~\cite{FCGF}}: \url{https://github.com/chrischoy/FCGF} (MIT License)

\noindent\textbf{DGR~\cite{DGR}}: \url{https://github.com/chrischoy/DeepGlobalRegistration} (MIT License)

\noindent\textbf{Minkowski Engine~\cite{minkowski}}: \url{https://github.com/NVIDIA/MinkowskiEngine} (MIT License)

\noindent\textbf{PointDSC~\cite{PointDSC}}: \url{https://github.com/XuyangBai/PointDSC}

\noindent\textbf{TEASER++~\cite{TEASER}}: \url{https://github.com/MIT-SPARK/TEASER-plusplus} (MIT License)

\noindent\textbf{Open3d~\cite{Open3D}}: \url{https://github.com/isl-org/Open3D} (MIT License)

\noindent\textbf{GC-RANSAC~\cite{GC_RANSAC}}: \url{https://github.com/danini/graph-cut-ransac} (new BSD License)

\noindent\textbf{Symmetric-ICP~\cite{SymICP}}: \url{https://gfx.cs.princeton.edu/proj/trimesh2/} (GPL Version 2 License)

\noindent\textbf{Best-Buddies Registration~\cite{BBR}}: \url{https://github.com/AmnonDrory/BestBuddiesRegistration}
\end{appendices}

\end{document}